\documentclass[10pt,twocolumn,letterpaper]{article}

\usepackage{iccv}
\usepackage{times}
\usepackage{epsfig}
\usepackage{graphicx}
\usepackage{amsmath}
\usepackage{amssymb}

\usepackage{multirow}
\usepackage{booktabs}

\usepackage[pagebackref=true,breaklinks=true,letterpaper=true,colorlinks,bookmarks=false]{hyperref}

\usepackage{subfiles}
\iccvfinalcopy % *** Uncomment this line for the final submission

\def\iccvPaperID{24} % *** Enter the ICCV Paper ID here

\ificcvfinal\fi

\begin{document}

\title{Weakly-Supervised Semantic Segmentation by Learning Label Uncertainty}

\author{Robby Neven, Davy Neven, Bert De Brabandere, Marc Proesmans and Toon Goedemé\\
ESAT-PSI, KU Leuven, Belgium\\
{\tt\small firstname.lastname@kuleuven.be}

}

\maketitle
\ificcvfinal\thispagestyle{empty}\fi

   \begin{abstract}
   Since the rise of deep learning, many computer vision tasks have seen significant advancements. However, the downside of deep learning is that it is very data-hungry. Especially for segmentation problems, training a deep neural net requires dense supervision in the form of pixel-perfect image labels, which are very costly. In this paper, we present a new loss function to train a segmentation network with only a small subset of pixel-perfect labels, but take the advantage of weakly-annotated training samples in the form of cheap bounding-box labels. Unlike recent works which make use of box-to-mask proposal generators, our loss trains the network to learn a label uncertainty within the bounding-box, which can be leveraged to perform online bootstrapping (i.e. transforming the boxes to segmentation masks), while training the network. We evaluated our method on binary segmentation tasks, as well as a multi-class segmentation task (CityScapes vehicles and persons). We trained each task on a dataset comprised of only 18\% pixel-perfect and 82\% bounding-box labels, and compared the results to a baseline model trained on a completely pixel-perfect dataset. For the binary segmentation tasks, our method achieves an IoU score which is ~98.33\% as good as our baseline model, while for the multi-class task, our method is 97.12\% as good as our baseline model (77.5 vs. 79.8 mIoU).
   \end{abstract}
\section{Introduction}
Nowadays, deep learning enables computer vision algorithms to achieve outstanding results for semantic segmentation tasks (\eg \cite{seg1}, \cite{seg2}, \cite{seg3}, \cite{seg4}). However, the standard deep learning approaches are fully supervised. This means they make use of a large labeled dataset to train the deep neural network. To construct such a dataset, one has to label each individual pixel to a specific class. Since most of the use-cases for segmentation tasks require images of high resolution (autonomous vehicles, medical imagery, ...), it is clear that this is a time-consuming job, making it one of the largest costs in deep learning.

Another way of labeling objects is by using bounding boxes (\ie, each object is enclosed by a rectangle). This way of labeling is extremely fast and cheap. Whereas for pixel-perfect labels a person has to label each individual pixel to a specific class, they now only have to mark two points of an enclosing rectangle and select the concerning class. Of course, bounding-box labels are not suitable to achieve good segmentation results right out of the box. Since the labels now contain a lot of erroneously labeled pixels, training a segmentation network with bounding-box labels will yield unexpected results. 

Recent works have already investigated this problem of training a segmentation model with weakly supervised data. Most of the works rely on box-to-mask proposal generator \cite{WSLL}, \cite{BCM}, \cite{SDI} to improve the bounding-box labels to better segmentation masks. However, since these generators are not specifically tuned to the dataset at hand, we believe that this method is not ideal. \cite{class-agnostic} also addressed this problem by learning a specific generator. However, they trained the mask generator on a different dataset, which is not always available.

In this paper, we propose to leverage label uncertainty to perform online bootstrapping on the bounding-box labels during training. Based on this learned label uncertainty, the online bootstrapping transforms the bounding boxes to more accurate label masks, eliminating the need to have generic mask generators. By feeding in a small portion of pixel-perfect labels, the model sees contradicting supervision around objects: background for pixel-perfect labels, while for the bounding-box labels the model sees the erroneous bounding-box foreground target around objects. This confusion can be used to learn a kind of label uncertainty on the fly, while only having a small portion of pixel-perfect labels. After only a few iterations, the model has learned to increase the uncertainty for erroneous bounding box targets. This can then be leveraged to bootstrap the bounding box labels to more accurate segmentation masks, by flipping them towards the model's output when the uncertainty exceeds a certain threshold. Due to the improving labels during training, the model's segmentation performance increases over time.

We tested our method on the CityScapes dataset for both binary and multi-class segmentation setups using bounding-box labels and a small portion of pixel-perfect labels and showed that our method is only slightly less accurate than training on a completely pixel-perfect dataset.

\section{Related Work}
In recent years, many researchers have investigated the problem of weakly-supervised semantic segmentation using bounding boxes. A lot of works follow the proposal generator methods. These techniques leverage proposal generators to generate, based on the given bounding-box data, enhanced segmentation masks for training the model. The most familiar proposal generation algorithms include dense CRF \cite{CRF-generator}, the GrabCut algorithm \cite{grabcut-generator} and the MCG algorithm \cite{MCG-generator}.

In the weakly supervised setting using bounding-boxes, most recent works are based on any of these proposal algorithms. For example, \cite{WSLL} uses the CRF algorithm to estimate segmentation proposals.  These proposals are then used as pseudo labels in the EM algorithm to optimize the DNN. Also, in \cite{BCM}, the authors introduced BCM, which exploits box-driven class-wise masking and class filling rates to remove irrelevant regions in the bounding boxes. However, their method also relies on the dense CRF proposal algorithm.

On the other hand, the BoxSup \cite{boxsup} and SDI \cite{SDI} algorithms rely on MCG to generate segmentation proposals. While the SDI also uses the GrabCut algorithm to generate more accurate proposals by combining them with the proposals generated by MCG. In Box2Seg \cite{box2seg}, the authors propose an additional DNN following the encoder-decoder architecture, which includes an attention module to further improve the segmentation performance.

In a recent paper \cite{class-agnostic}, the authors suggested to instead of using the traditional proposal algorithms, they learn a pseudo-mask generator on a different dataset and use this generator to create masks for the given bounding-boxes to improve the semantic segmentation.

While all the discussed works rely on the generation of mask proposals, we suggest a proposal-free method. As most works go through different iteration stages to improve the labels and fine-tune the segmentation model, our method trains the model without having to statically update labels or refine-tune the model, since the box-to-mask conversion is built-in into our loss function and does not introduce extra overhead (in contrast to mask proposal generators). 
\section{Method}
This section will mainly focus on our loss function by describing the two main parts it consists of. First, we will explain the uncertainty loss, which enables the model to learn the label uncertainty, indicating where the target label might be incorrect. Secondly, we will describe how the learned uncertainty can be used to perform online bootstrapping, which will, during the training phase, adapt the erroneous targets on the fly to improve the training supervision and increase the model's performance.

\subsection{Learning Uncertainty}

An early work by Kendall and Gal \cite{kendall2017uncertainties} modeled the uncertainty by replacing the output layer with a Gaussian distribution. By replacing the output logit with an output mean $\mu$ and variance $\sigma^2$, the model can indicate its uncertainty by increasing the latter. For example, if we look at a standard regression case using an L2-loss, the objective is to minimize the squared error between the inferred value and the label y. If the model now instead infers a Gaussian distribution with mean $\mu$ and variance $\sigma^2$, the objective changes to maximizing the target's probability w.r.t. this inferred distribution, that is, maximizing $P(y | \mu, \sigma^2)$ with  $P \sim \mathcal{N}(\mu,\,\sigma^{2})$.

In \cite{kendall2017uncertainties}, Kendall and Gal argued that this uncertainty represents the heteroscedastic uncertainty of the model. This type of uncertainty captures noise inherent to the model's input (\eg blurred regions, poor lighting, occlusions, etc.), which cannot be explained away using more training data. Another advantage of using this method to learn the heteroscedastic uncertainty is that it is learned completely unsupervised.

If one looks more closely at the new objective of maximizing $P(y | \mu, \sigma^2)$ with  $P \sim \mathcal{N}(\mu,\,\sigma^{2})$ in Equation \ref{eq:gaussian_likelihood} , the loss can be rearranged to a simpler form, which gives more insight in the uncertainty's influence (Equation \ref{eq:regression_loss_w_unc}).

\begin{equation}
    \label{eq:gaussian_likelihood}
    P(y | \mu, \sigma^2) = \frac{1}{\sigma\sqrt{2\pi}}e^{-\frac{1}{2}(\frac{y-\mu}{\sigma})^2}
\end{equation}
Which leads to minimising:
\begin{equation}
    \label{eq:regression_loss_w_unc}
    -\log P(y| \mu, \sigma^2) = 
    \frac{1}{2\sigma^2}(y - \mu)^2 + \frac{1}{2}\log{\sigma^2}
\end{equation}

\begin{figure}
\begin{center}
   \includegraphics[width=0.9\linewidth]{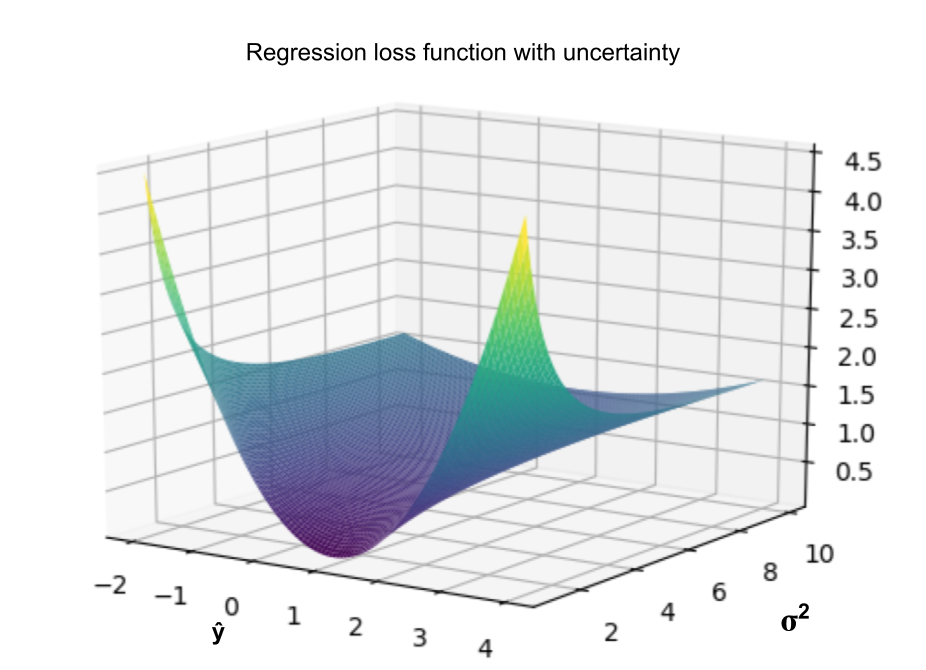}
\end{center}
   \caption{Regression loss with uncertainty for target value 0. It is clear the loss is minimized when outputting a mean value of 0. However, by increasing the uncertainty, the loss is also lowered for other mean values which deviate from 0.}
\label{fig:uncertainy_regression}
\end{figure}

It is clear the rearranged form now consists of an L2-loss attenuated by $\frac{1}{2\sigma^2}$ which is also penalized by an additional uncertainty term $\frac{1}{2}\log\sigma^2$. By increasing the uncertainty, the model can attenuate the loss but is also penalized with an increasing penalty, prohibiting the model to minimize the loss indefinitely by increasing the uncertainty for all pixels. An overview of the loss' curvature can be seen in Figure \ref{fig:uncertainy_regression}.

\begin{figure}
    \begin{center}
   \includegraphics[width=\linewidth]{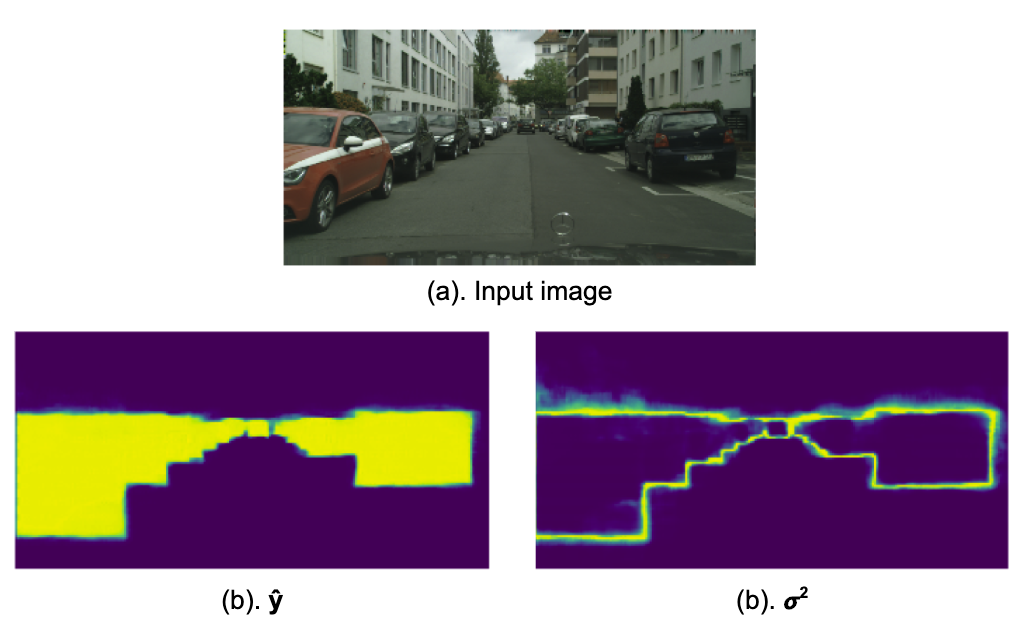}
\end{center}
    \caption{Output of a segmentation model trained on bounding box-data using the regression loss with uncertainty. It is clear that, when the dataset only consists of bounding-box targets, the segmentation output will also resemble rectangles.}
    \label{fig:unc_example}
\end{figure}

As explained above, this loss models the heteroscedastic uncertainty, enabling the model to express its uncertainty for blurred, poorly lit, or difficult regions in images. For these difficult pixels, the model can increase its uncertainty and infer a mean value that contradicts the target label, since a high uncertainty also decreases the loss (Equation~\ref{eq:regression_loss_w_unc}). When training a binary segmentation model using this loss function with bounding-box supervision, the most difficult pixels to learn will be around the edges of the boxes. Therefore, the uncertainty will be high, and the model will infer a more relaxed bounding box since not all bounding boxes in the training data are strictly enclosing objects. Figure \ref{fig:unc_example} shows the output of such a scenario. The model learns to segment bounding boxes, but the uncertainty is high at the boxes' edges.

We already stated that the uncertainty is learned unsupervised. The way the model learns this uncertainty is that for the many examples in the training set, the bounding box circumference of the object is not always exactly the same. Therefore, the model sees a variance in the width and height of the bounding boxes during training and will contradict training targets for the pixels at the bounding boxes' borders. Hence, the model will learn a generalized bounding box shape and contradict the target pixels outside of the bounding box by increasing the uncertainty to lower the loss function.

\subsection{Confusing the model with pixel-perfect labels.}
Our goal is to perform semantic segmentation, which can segment objects with their true borders instead of rectangular shapes as seen in Figure~\ref{fig:unc_example}. Instead of only learning the uncertainty for "difficult" pixels, which is built-in in the loss function, we will force the loss to learn a kind of label uncertainty. This new kind of uncertainty will give us an indication of where the training label might be incorrect, and will therefore enable the model to lower the loss (through the attenuation term from Equation \ref{eq:regression_loss_w_unc}) when contradicting the target label. We already showed that the model learns its uncertainty for difficult regions, but also regions where the model gets contradicting target labels during training (\eg edges which are coarsely labeled). Therefore, if we feed in a small number of pixel-perfect labels during training, where objects' borders are pixel perfectly annotated, the model will get confused for regions around objects with correctly annotated background targets (pixel-perfect labels), and erroneous foreground targets (bounding-box labels). This contradiction in supervision enables the loss function to correctly learn the label uncertainty.

To illustrate this idea, we will look at a binary car segmentation model. The model infers a Gaussian distribution, that is, two output maps: mean $\mu$ and variance $\sigma^2$. Consider a training batch, which consists of both pixel-perfect (PP) and bounding-box (BBox) targets. The loss for the pixel-perfect labels is a standard L2 loss with treating the inferred mean value as the standard logit output. The loss for the bounding box labels is the extended L2 loss with uncertainty (Equation \ref{eq:regression_loss_w_unc}). Eventually, to compute the total batch loss, the pixel's losses are accumulated. However, for pixels around objects, the model sees contradicting supervision: foreground (target y = 1) for bounding-box targets, background (target y = 0) for pixel-perfect targets.  

\begin{equation}
\label{eq:summed_loss_reg}
\begin{split}
    \mathcal{L}  & = \mathcal{L}(PP) + \mathcal{L}(BBox) \\
    & = (y - \mu)^2 + \frac{1}{2\sigma^2} (y - \mu)^2 + \frac{1}{2}\log{\sigma^2} \\
    \intertext{Substituting y = 0 for pixel-perfect labels and y = 1 for bounding box labels}
    \mathcal{L} & = (0 - \mu)^2 + \frac{1}{2\sigma^2} (1 - \mu)^2 + \frac{1}{2}\log{\sigma^2}
\end{split}
\end{equation}

\begin{figure}
    \begin{center}
   \includegraphics[width=\linewidth]{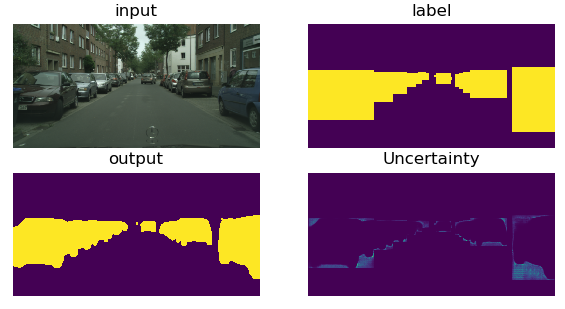}
\end{center}
    \caption{Output example of a binary segmentation model trained with an L2-loss with uncertainty using a small subset of pixel-perfect labels. The pixel-perfect labels enable the model to learn to increase the uncertainty for the erroneous bounding-box pixels. For these pixels, the uncertainty attenuation term lowers the loss, enabling the model to contradict the erroneous bounding-box supervision which results in a finer segmentation output.}
    \label{fig:regression_with_uncertainty_output}
\end{figure}

We see in Equation \ref{eq:summed_loss_reg} that the first term (L2-loss for pixel-perfect targets) is minimized by inferring the same $\mu$ as the target label (0, background). However, if the model learns to infer a 0 for background pixels around objects for the pixel-perfect labels, it will also infer this value for images supervised with bounding-box targets (foreground). If we look at Equation \ref{eq:summed_loss_reg}, when inferring a mean ($\mu$) value of 0 (model infers background), the loss for the bounding-box targets can only be minimized by increasing the uncertainty $\sigma^2$. Hence, due to a small subset of pixel-perfect labels, the model can learn the label uncertainty in an unsupervised manner. An example of this method can be seen in Figure \ref{fig:regression_with_uncertainty_output}. The figure clearly shows the learned uncertainty for the erroneous bounding-box targets, lowering the loss through the attenuation term which enables the model to infer the contradicting background class resulting in finer segmentation masks.

\subsubsection{Switching to the Cross-Entropy Loss}
In the previous sections, we discussed how the use of label uncertainty can help train a segmentation model using bounding-box labels. To illustrate the effect, we used the simple L2-loss to clearly see the attenuation and penalty effect on the loss when the uncertainty increases. However, a standard binary segmentation task is nearly always trained with a binary cross-entropy loss (BCE), and will usually result in better performance.

When using the BCE loss, the last layer of the network outputs a logit value $l$, which is the logarithm of the odd ($\log(\frac{p}{1-p})$). The logit can range from $-\infty$ to $+ \infty$. It gets squashed through the sigmoid function (eq. \ref{eq:sigmoid}) to get a value between 0 and 1, representing the probability $\hat{p}$ of that pixel being 1 (which can be derived by inserting the logarithm of the odd into the sigmoid). Next, the BCE loss function (Equation \ref{eq:BCE}) measures the cross-entropy between the probability of the network's output $\hat{p}$ and the probability of the label $p$.

\begin{equation}
    \label{eq:sigmoid}
     \hat{p} = sigmoid(l) = \frac{1}{1+e^{-l}}
\end{equation}

\begin{equation}
    \label{eq:BCE}
    \mathcal{L}(\hat{p}, p) = p\log \hat{p} + (1-p)\log (1-\hat{p}) 
\end{equation}

To add uncertainty to the BCE loss, we refer again to the implementation of Kendal and Gal \cite{kendall2017uncertainties}. Like in the previous sections, the model will, instead of outputting a single logit value, output a Gaussian distribution of logits with mean $\mu$ and variance $\sigma^2$. Whereas in the standard BCE loss, the logit gets squashed through the sigmoid to get the probability $\hat{p}$, we have to compute E[$\hat{p}$], the expected value of the probability:

\begin{equation}
    E[\hat{p}(l)] = \int sigmoid(l)\cdot P(l)dl
\end{equation}
With P(l):
\begin{equation}
    P(l) = \frac{1}{\sigma\sqrt{2\pi}}\exp^{-\frac{1}{2}(\frac{l-\mu}{\sigma})^2}
\end{equation}

Solving the integral to get the expected probability can be done by using Monte Carlo integration:  logit values get sampled from the Gaussian distribution outputted by the network, squashed through the sigmoid function and averaged. Luckily, \cite{xiaopeng} approximated the integral:

\begin{equation}
    \label{expected_sigmoid}
    E[\hat{p}(l)] = sigmoid(\frac{\mu}{\sqrt{1+\pi\sigma^2/8}}) 
\end{equation}

The expected probability can be computed by first computing an "adapted" logit value from $\mu$ and $\sigma^2$ and taking the sigmoid function to get E[$\hat{p}$]. The loss value is the standard BCE of E[$\hat{p}$]:

\begin{equation}
    \label{eq:BCE_WITH_UNC}
     \mathcal{L}(\hat{p}, p) = p\log E[\hat{p}] + (1-p)\log (1-E[\hat{p}])
\end{equation}

\begin{figure*}[tb]
    \begin{center}
    \includegraphics[width=\linewidth]{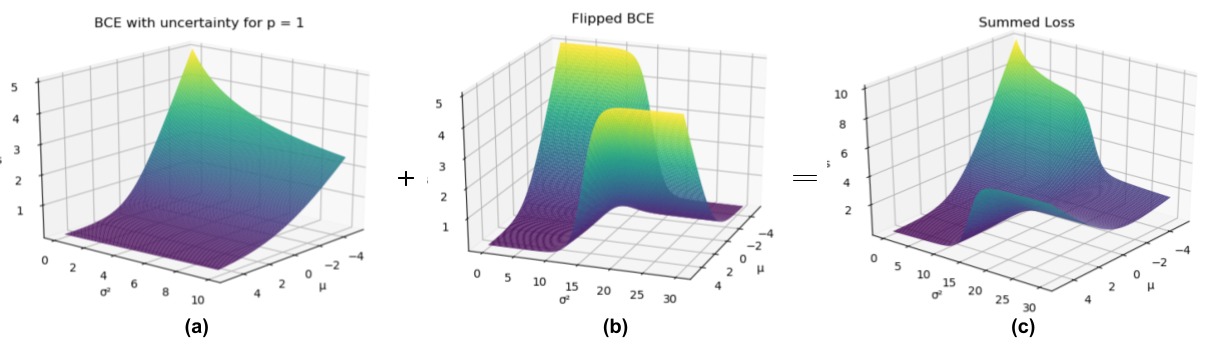}
    \end{center}
    \caption{\textbf{(a)} BCE loss function with uncertainty for target value 1. It is clear the loss can be lowered by increasing the uncertainty $\sigma^2$. However, the loss can be minimized this way only up to a certain point. Also, the loss still has significant gradients to shift mean $\mu$ towards the target value 1. For erroneous background pixels within bounding-box targets, this means that the model will learn to segment rectangular shapes if no extra pixel-perfect labels are used during training. \textbf{(b)} Bootstrapping loss. The loss is the sum of two BCE losses: the first receives the original target while the second receives the flipped target (opposite). The two terms are weighed based on the variance $\sigma^2$. Higher variance means the loss consists of the BCE with the flipped target. For low variance, the loss resembles the original BCE with the given training target. \textbf{(c)} The total loss function, which consists of two summed loss terms. However, the variance (uncertainty) is only learned in the uncertainty loss, while the mean (logit) is learned in the bootstrapping loss. }
    \label{fig:loss_function}
\end{figure*}

Again, the uncertainty term acts as an attenuation of the loss, enabling the model to lower the loss even when inferring a contradicting target. However, the same problem as with the L2-loss arises. In Figure \ref{fig:loss_function} (a), it is clear the loss decreases when the uncertainty increases, but at each point, gradients are directing the mean towards the target label. For background pixels within bounding-box targets, this means the model will still learn to segment rectangular shapes. However, we have addressed this problem in the previous section by confusing the model with pixel-perfect labels, which will prevent the model from learning from these "false" gradients. 

\subsection{Uncertainty-Based Online Bootstrapping}
In the previous sections, we showed how the uncertainty loss could be used to train a pixel-perfect segmentation model with bounding-box data. However, the downside was that we needed to use pixel-perfect labels, to mitigate the gradients towards the erroneous training targets within the bounding box. Still, when following this method, the model will try to minimize the loss on background pixels within bounding boxes by increasing the uncertainty, but the loss function will remain to provide gradients towards the training target y.

To eliminate these gradients, we propose to add online bootstrapping directly into the loss function. We already stated that the uncertainty represents a label uncertainty, that is, a degree of how certain the model is that the target is incorrect. Having this information, we can flip the training target to background for every pixel where the uncertainty exceeds a certain threshold. By flipping the target to background, the "false" gradients towards old targets $y$ are replaced with gradients towards the flipped target $y^*$. However, we have to keep in mind the uncertainty is learned on $y$. Therefore, we compose the loss function of two parts: the uncertainty loss and a bootstrapping loss.

The bootstrapping loss is a standard BCE loss with the bootstrapped target. This target can be the training target $y$, or, when the uncertainty reaches a predetermined threshold value $\tau$, the flipped training target $y^*$. To achieve a smooth, differentiable loss function, we compose the bootstrapping loss (Equation \ref{eq:bootstrapping_loss}) of two summed terms: a BCE with the training target $y$ and a BCE with the flipped target $y^*$, which are weighted based on the uncertainty $\sigma^2$ (Figure \ref{fig:loss_function} (b)).

\begin{equation}
    \mathcal{L} = W\cdot BCE(\mu, y) + (1 - W)\cdot BCE(\mu, y^*)
    \label{eq:bootstrapping_loss}
\end{equation}

With

\begin{equation}
    W = sigmoid\left(\frac{\tau - \sigma^2}{0.2}\right)
\end{equation}

As we have already pointed out, for an increased uncertainty (i.e. the model contradicts the training target), the uncertainty loss still has strong gradients towards the training target. However, by adding the pixel-perfect labels to each batch, we forced the model to not follow these gradients, since this would undermine the performance on the pixel-perfect labels. This situation is not ideal, because the gradients towards the training targets limit the training. 

By adding the bootstrapping loss, we introduce the extra gradients towards the flipped targets (when the uncertainty is high). This enables the model to receive the gradients it prefers. However, these flipped gradients only happen when the model has increased its uncertainty through the uncertainty loss. These two learning mechanics can be split between the two different loss terms. Learning the uncertainty $\sigma^2$ is restricted to the uncertainty loss while learning the mean $\mu$ is handled by the bootstrapping loss. This is easily done in the PyTorch framework \cite{PyTorch}, by detaching the mean or variance for each of the corresponding loss terms. Figure \ref{fig:loss_function} (c) shows the final curvature of our uncertainty-based online bootstrapping loss.

\section{Experiments \& Results}
In this section, we will show the performance of our new loss function on a binary and multi-class setup. First, we will perform binary segmentation of the \textit{cars} class from the CityScapes dataset \cite{Cordts2016CityScapes}, comparing both the L2 with uncertainty (without bootstrapping) and the BCE uncertainty loss function with bootstrapping. In a second experiment, we will extend our loss function to a multi-class setting, combining the \textit{cars and persons} classes into one dataset.

\subsection{Segmenting Cars from CityScapes}
To test our method, we will first perform binary segmentation on from the CityScapes dataset. We will use the 3475 available train and validation images, and divide them into an 80:20 train/validation split. To speed up our experiments, the images are scaled to a resolution of 1024$\times$512. Furthermore, we will use the PyTorch deep learning framework \cite{PyTorch} to train a standard ERFNet segmentation network \cite{ERFNet} with an Adam optimizer \cite{adam} and a learning rate of $5\cdot 10^{-4}$. Since we only focus on the performance of our loss function, we will not optimize the ideal network architecture and learning rate optimization.

\subsubsection{Regression with uncertainty}
In the first experiment, we will show the performance of the regression loss with uncertainty without bootstrapping. For this setup, we will use a fixed set of 500 pixel-perfect labels together with the remaining 2280 bounding-box labels. A batch size of 8 is used, with a pixel-perfect sampling chance of 25\%, to make sure that each batch has around two pixel-perfect examples. We compare training the uncertainty on all pixels and training the uncertainty only on foreground pixels (standard L2 loss for background pixels). Table \ref{tab:reg_results} gives an overview of the results. Compared to the baseline using a standard BCE loss training on 500 pixel-perfect labels (78.6\% IoU), we see that, when training together with the bounding-box labels, the uncertainty loss manages to increase the IoU to 79.02\% when training the uncertainty on all pixels and 83.04\% when only training on foreground pixels. Whereas a standard BCE loss achieves an IoU score of only 69.43\% when training on the same mixed set of pixel-perfect and bounding-box labels.

% \usepackage{multirow}
\begin{table}
\begin{center}
\begin{tabular}{c|cc|c}
\hline
\textbf{Loss}                                                                         & \multicolumn{2}{c|}{\textbf{Dataset}} & \textbf{IoU (\%)} \\ \hline
                                                                                      & \textbf{PP}       & \textbf{BB}       & \textbf{Car}  \\ \hline 
\multirow{3}{*}{{Standard BCE}}                                                & 2780              & /                 & 86.9              \\
                                                                                      & 500               & /                 & 78.6              \\
                                                                                      & 500               & 2780              & 69.43             \\ \hline
{Regression w/ Uncertainty}                                                             & 500               & 2280              & 79.02             \\ \hline
{\begin{tabular}[c]{@{}c@{}}Regression w/ Uncertainty\\ (Foreground Only)\end{tabular}} & 500               & 2280              & 83.04             \\ \hline
\end{tabular}
\end{center}
\caption{Results for the regression loss function with uncertainty. We tested this loss on the CityScapes car segmentation task with a 500:2780 ratio pixel-perfect to bounding-box labels. We compare against the baselines of using all pixel-perfect and only 500 pixel-perfect labels with a standard BCE loss. The bottom two rows compare the loss function when training the uncertainty on all pixels (second last row), and when only training the uncertainty within the bounding box (last row).}
\label{tab:reg_results}
\end{table}

\subsubsection{BCE with Uncertainty \& Online Bootstrapping}
\begin{figure*}
    \begin{center}
    \includegraphics[width=\linewidth]{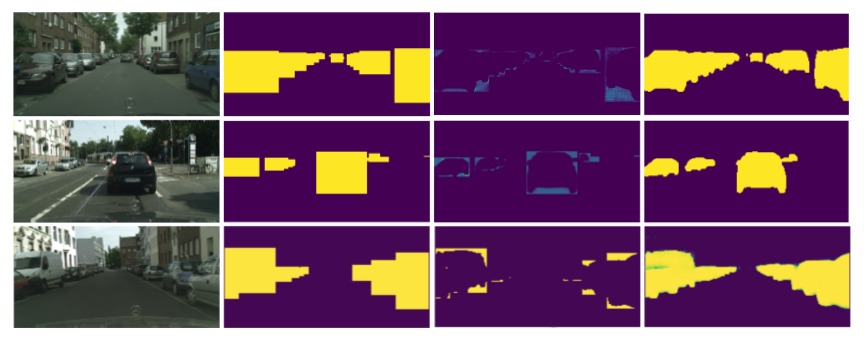}
    \end{center}
    \caption{Training output using the BCE w/ uncertainty loss together with the online bootstrapping. From left to right: input image, training target, uncertainty weight mask (higher means target gets flipped to background), and segmentation output. The top two rows correspond from batches earlier during training, the last row is captured near the end of training. It is clear the model learns the label uncertainty significantly well.}
    \label{fig:bce_results}
\end{figure*}

To improve our loss function further, we will replace the regression loss with the BCE loss as described in the method section. We will also insert the online bootstrapping with a fixed threshold of 2.5. Since in our previous experiment we concluded that the learned uncertainty is better when only trained within the bounding boxes, we will again only apply our loss within the bounding box and apply a standard BCE to background pixels. To see the influence of the number of pixel-perfect labels, we will train the model on a varying ratio of 500:2280 to 100:2680 pixel-perfect to bounding-box labels. Table \ref{tab:classification_results} gives an overview of the results. Figure \ref{fig:bce_results} shows some training examples, including the bootstrapping weight masks.

\begin{table}
\begin{center}

\begin{tabular}{c|cc|c}
\hline
\textbf{}           \textbf{Loss}                                                    & \multicolumn{2}{c|}{\textbf{Dataset}} & \textbf{IoU (\%)} \\ \hline
                                                                          & \textbf{PP}                & \textbf{BB}             &  \textbf{Car}  \\ \hline 
\textbf{Standard BCE}                                                     & 2780              & /                & 86.9              \\
                                                                          & 500               & /                & 78.6              \\
                                                                          & /                 & 2780             & 69.18             \\
                                                                          & 500               & 2280             & \textbf{69.43}             \\ \hline
\textbf{\begin{tabular}[c]{@{}l@{}}BCE Loss With \\ Uncertainty\end{tabular}} & 500               & 2280             & \textbf{85.45}             \\
                                                                          & 400               & 2380             & 85.32             \\
                                                                          & 300               & 2480             & 85.42             \\
                                                                          & 200               & 2580             & 84.37             \\
                                                                          & 100               & 2680             & 81*               \\ \hline
\end{tabular}
\end{center}
\caption[Results of binary segmentation using classification loss]{The table shows the results for both the standard BCE loss and the classification loss with uncertainty developed in this chapter. (*) The output of this experiment resembles rectangles and is not the expected segmentation result.}
\label{tab:classification_results}
\end{table}

The best performance we can get using 500 pixel-perfect labels is an IoU score of 85.45\%, which is around 1.5\% lower than the baseline using 2780 pixel-perfect labels. This indicates our method can successfully achieve near-perfect segmentation results, using only a fraction of pixel-perfect data. Even if the amount of pixel-perfect labels is reduced to only 200, the IoU does not drop significantly, meaning our loss function does not need to have that many pixel-perfect examples to learn the label uncertainty. However, we noticed that when using a pixel-perfect set of 100 images, the segmentation output converged to the bounding-box targets.

\subsection{Extending to Multi-Class}
In the previous experiments, we showed that our method works significantly well on binary segmentation. However, ideally, we want our method to work on multi-class segmentation problems as well. Therefore, we extended our loss function to suit a multi-class problem by changing two main things. First, we substituted the \textit{BCE with uncertainty} term with a \textit{Cross-entropy loss with uncertainty} as implemented by \cite{kendall2017uncertainties}. In short, instead of outputting a logit distribution with mean $\mu$ and variance $\sigma^2$: $logit \sim \mathcal{N}(\mu,\,\sigma^{2})$, the network now outputs a multi-variate Gaussian logit distribution with a mean vector of length C and a diagonal covariance matrix $\Sigma$, with for each class its corresponding variance $\sigma^{2}_c$ so that the logit vector now follows the distribution $\textbf{l }\sim \mathcal{N}(\boldsymbol{\mu},\,\Sigma)$.
Normally, the logit vector gets transformed by a softmax function to a probability distribution $\hat{p}$. However, since we now have a distribution of logit vectors, we have to compute the expected probability distribution $E[\hat{p}]$. Next, to compute the loss, we have to take the cross-entropy loss of $E[\hat{p}]$:

\begin{equation}
    \label{eq:multiclass_loss}
    \mathcal{L} = -p_y\log(E[\hat{p}_y])
\end{equation}

With for $E[\hat{p}]$

\begin{equation}
    E[\boldsymbol{\hat{p}}] = \int softmax(\boldsymbol{l})\cdot P(\boldsymbol{l})d\boldsymbol{l}
\end{equation}

This integral is not able to be computed with a closed-form solution and can only be estimated through MCI, by sampling logit vectors from the output distribution. \cite{kendall2017uncertainties} proposed the numerically stable solution:

\begin{figure*}[htb!]
    \begin{center}
    \includegraphics[width=0.9\linewidth]{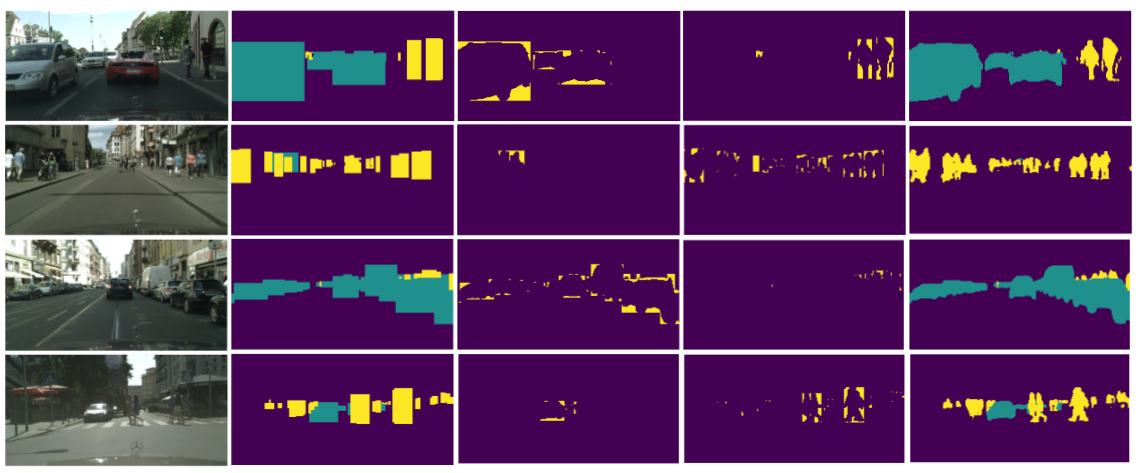}
    \end{center}
    \caption{Training output of the multi-class network. From left to right: input image, target masks, car flip mask, person flip mask, segmentation output. The flip masks indicate within the bounding boxes which pixels have high uncertainty, and receive therefore a flipped training target (i.e. the output of the model at those pixels).}
    \label{fig:multi-class}
\end{figure*}

\begin{equation}
    \boldsymbol{l_t} = \boldsymbol{\mu} + \boldsymbol{\Sigma} \odot\boldsymbol{\epsilon_t},\  \boldsymbol{\epsilon_t} \sim \mathcal{N}(\boldsymbol{\mu},\,\boldsymbol{\Sigma})
\end{equation}
\begin{equation}
    \mathcal{L}_x = \log\frac{1}{T}\sum_{t}\exp(l_{t, c} - \log\sum_{c'}\exp l_{t, c'})
\end{equation}

When trained with a mixed dataset of PP- and BB-labels, the confusion in supervision near objects will again cause the uncertainty to resemble the label uncertainty. When the label uncertainty is high, certain classes exhibit high variance which will, in turn, lower the loss, and enables the model to infer a different class, contradicting the target.

Where in the binary case the second term of our loss function (i.e. the bootstrapping term) was a sum of $BCE(\mu, y)$ and $BCE(\mu, y^*)$, it now represents a single term, that is, the cross-entropy loss w.r.t. the bootstrapped target: $CE(\boldsymbol\mu, y*)$. We know that the label uncertainty is high when some classes exhibit high uncertainty. Therefore, if one class exhibits a high variance above a predetermined threshold $\tau$, then we flip the target $y$ to the model's output class (i.e. the class with the highest mean score). 

\begin{equation}
    y = \begin{cases}
         \text{argmax}(\boldsymbol{\mu}) & \text{ if } \text{max}(\boldsymbol{\sigma^2}) > \tau \\ 
         y & \text{ if } \text{max}(\boldsymbol{\sigma^2}) < \tau
        \end{cases}
\end{equation}

We tested this loss on the CityScapes dataset, segmenting the person and car instances. Table \ref{tab:multi-class} shows a summary of the results. Figure \ref{fig:multi-class} shows some example images during training with their corresponding bootstrapped target masks. The proposed loss function works significantly well, scoring a mean IoU which is only ~2\% IoU lower than the baseline using only PP-labels. However, due to the randomness of the loss function (MCI and setting a suitable threshold value), we experience numerical instabilities when extending the loss to more than 3 classes. This will be investigated in future work.

\begin{table}
\begin{center}
\begin{tabular}{c|cc|cccc} \toprule
\textbf{Loss}       & \multicolumn{2}{c|}{\textbf{Dataset}}       & \multicolumn{4}{c}{\textbf{IoU (\%)}}                                                            \\ \hline
\textit{\textbf{}}  & \textit{\textbf{PP}} & \textit{\textbf{BB}} & \textit{\textbf{Mean}} & \textit{\textbf{BG}} & \textit{\textbf{Car}} & \textit{\textbf{Person}} \\
\multirow{2}{*}{CE} & 2780                 & /                    & 79.80                  & 98.21                & 85.82                 & 55.68                    \\
                    & 500                  & /                    & 73.36                  & 97.55                & 79.94                 & 42.58                    \\
Ours                & 500                  & 2280                 & 77.75                  & 98.01                & 82.05                 & 53.2         \\           \bottomrule
\end{tabular}
\end{center}
\caption[Multi-class segmentation results]{Multi-class segmentation results. The first two rows show the cross-entropy loss baseline IoUs for training sets comprising of 2780 and 500 pixel-perfect labeled images respectively. The third row shows the result of a model trained with our new loss function on a training set consisting of 500 ground truth and 2280 bounding box images. Training the model on the mixed PP/BB dataset using a standard CE loss results in rectangular segmentation output.}
\label{tab:multi-class}
\end{table}

\section{Discussion}
As we have shown in the previous section, the models trained with our loss function can achieve nearly the same results as a fully pixel-perfect dataset with now only ~18\% pixel-perfect labels. However, for the binary segmentation case, we have shown that this number can be reduced with only a small drop in IoU. It is also important to remark the significance of the uncertainty threshold $\tau$ used for bootstrapping. We have examined that when this threshold is too small, many foreground pixels get flipped to background, which disables further learning. However, after finding the right setting, we haven't seen this issue anymore, but we expect that it is different for other segmentation problems or datasets.

Also, we have to remark that for the multi-class setting, the loss function uses the Monte Carlo Integration technique to calculate the uncertainty loss. This introduces randomness to the loss and is dependent on the number of samples taken. During training, this sampling does not introduce significant overhead, since we could get adequate results with only 20 samples per batch. Taking more samples will increase the computational overhead, but we are not sure if this would improve the segmentation performance. However, when extending the multi-class loss to more than three classes, we experienced numeric instabilities which prevented us from training the model on more classes. We will investigate this in future work.
\section{Conclusion}
In this paper, we proposed a method for training a segmentation model with bounding box labels without the need of box-to-mask proposal generators. Our method requires only a small subset of pixel-perfect labels, which drastically reduces the annotation cost. The small number of pixel-perfect labels are used to efficiently learn label uncertainty for the bounding box labels. With this learned label uncertainty, the bounding-box labels can be updated during training to perform online bootstrapping, that is, transforming the boxes into pixel-perfect segmentation masks. Since these masks continuously improve, so does the segmentation performance. We have shown the performance of our method on the binary and multi-class setups. However, due to the sampling nature of the MCI in our loss function, extending our method to more than 3 classes is still work in progress.

\paragraph{Acknowledgment}
This research received funding from the Flemish Government (AI Research Program).

{\small
\bibliographystyle{ieee_fullname}
\bibliography{main}
}

\end{document}